\documentclass{article}

\usepackage{microtype}
\usepackage{graphicx}
\usepackage{subfigure}
\usepackage{booktabs} 

\usepackage{hyperref}
\usepackage{enumitem}

\usepackage[accepted]{icml2024}

\usepackage{amsmath}
\usepackage{amssymb}
\usepackage{mathtools}
\usepackage{amsthm}

\usepackage[capitalize,noabbrev]{cleveref}

\theoremstyle{plain}

\theoremstyle{definition}

\theoremstyle{remark}

\usepackage[textsize=tiny]{todonotes}

\usepackage{multirow}

\icmltitlerunning{scTransNet for GRN Inference from Pre-trained scRNA-seq Transformer with Joint Graph Learning}

\begin{document}

\twocolumn[
\icmltitle{Gene Regulatory Network Inference from Pre-trained Single-Cell Transcriptomics Transformer with Joint Graph Learning}



\icmlsetsymbol{equal}{*}

\begin{icmlauthorlist}
\icmlauthor{Sindhura Kommu}{vtb}
\icmlauthor{Yizhi Wang}{vta}
\icmlauthor{Yue Wang}{vta}
\icmlauthor{Xuan Wang}{vtb}
\end{icmlauthorlist}

\icmlaffiliation{vtb}{Department of Computer Science, Virginia Polytechnic Institute and State University, Blacksburg, VA, United States}
\icmlaffiliation{vta}{Department of Electrical and Computer Engineering, Virginia Polytechnic Institute and State University, Arlington, VA, United States}

\icmlcorrespondingauthor{Sindhura Kommu}{sindhura@vt.edu}
\icmlcorrespondingauthor{Yizhi Wang}{yzwang@vt.edu}
\icmlcorrespondingauthor{Yue Wang}{yuewang@vt.edu}
\icmlcorrespondingauthor{Xuan Wang}{xuanw@vt.edu}

\icmlkeywords{transformer, single cell, gene regulatory networks (GRNs), scRNA-seq, inference, graph neural networks}

\vskip 0.3in
]



\printAffiliationsAndNotice{}  

\begin{abstract}
Inferring gene regulatory networks (GRNs) from single-cell RNA sequencing (scRNA-seq) data is a complex challenge that requires capturing the intricate relationships between genes and their regulatory interactions. In this study, we tackle this challenge by leveraging the single-cell BERT-based pre-trained transformer model (scBERT), trained on extensive unlabeled scRNA-seq data, to augment structured biological knowledge from existing GRNs. We introduce a novel joint graph learning approach scTransNet that combines the rich contextual representations learned by pre-trained single-cell language models with the structured knowledge encoded in GRNs using graph neural networks (GNNs). By integrating these two modalities, our approach effectively reasons over both the gene expression level constraints provided by the scRNA-seq data and the structured biological knowledge inherent in GRNs. We evaluate scTransNet on human cell benchmark datasets from the BEELINE study with cell type-specific ground truth networks. The results demonstrate superior performance over current state-of-the-art baselines, offering a deeper understanding of cellular regulatory mechanisms.

\end{abstract}

\section{Introduction}




Single-cell RNA sequencing (scRNA-seq) has transformed the exploration of gene expression patterns at the individual cell level \cite{Jovic2022-gq}, offering an unprecedented opportunity to unravel the intricate regulatory mechanisms governing cellular identity and function \cite{Pratapa2020}. One such promising application is the inference of gene regulatory networks (GRNs) which represent the complex interplay between transcription factors (TFs) and their downstream target genes \cite{akers2021gene, cramer2019organization}. A precise understanding of GRNs is crucial for understanding cellular processes, molecular functions, and ultimately, developing effective therapeutic interventions \cite{Biswas2021-ud}.

However, inferring GRNs from scRNA-seq data is challenging due to cell heterogeneity \cite{wagner2016revealing}, cell cycle effects \cite{buettner2015computational}, and high sparsity caused by dropout events \cite{kharchenko2014bayesian}, which can impact accuracy and robustness. Additionally, the availability of labeled scRNA-seq data corresponding to a GRN is limited, making it challenging to train models from scratch. Traditional unsupervised or self-supervised models, while not reliant on label information, often struggle to effectively handle the noise, dropouts, high sparsity, and high dimensionality characteristics of scRNA-seq data \cite{moerman2019grnboost2, matsumoto2017scode, zeng2023inferring}. Supervised methods are also proposed for GRN reconstruction \cite{zhao2022hybrid, shu2022boosting, KC2019, 10.1093/bioinformatics/btac559} but struggle to handle batch effects and fail to leverage latent gene-gene interaction information effectively limiting their generalization capabilities.

Recent advancements in large language models (LLMs) and the pre-training followed by fine-tuning paradigm \cite{devlin2019bert, openai2023gpt} have significantly contributed to the development of transformer-based architectures tailored for scRNA-seq data analysis \cite{Yang2022, Cui2024, Chen2023, Theodoris2023}. These models effectively leverage vast amounts of unlabeled scRNA-seq data to learn contextual representations and capture intricate latent interactions between genes. To address the limitations of the current methods, we effectively leverage one of these large-scale pre-trained transformer models, namely scBERT \cite{Yang2022}, which has been pre-trained on large-scale unlabelled scRNA-seq data to learn domain-irrelevant gene expression patterns and interactions from the whole genome expression. By fine-tuning scBERT on user specific scRNA-seq datasets, we can mitigate batch effects and capture latent gene-gene interactions for downstream tasks.

We propose an innovative knowledge-aware supervised GRN inference framework, scTransNet (see \cref{overall_framework}), which integrates pre-trained single-cell language models with structured knowledge of GRNs. Our approach combines gene representations learned from scBERT with graph representations derived from the corresponding GRNs, creating a unified context-aware and knowledge-aware representation \cite{feng2020scalable}. This joint learning approach enables us to surpass the accuracy of current state-of-the-art methods in supervised GRN inference. By harnessing the power of pre-trained transformer models and incorporating biological knowledge from diverse data sources, such as gene expression data and gene regulatory networks, our approach paves the way for more precise and robust GRN inference. Ultimately, this methodology offers deeper insights into cellular regulatory mechanisms, advancing our understanding of gene regulation.

\section{Related Work}
Several methods have been developed to infer GRNs from scRNA-seq data, broadly categorized into unsupervised and supervised methods.

\textbf{Unsupervised methods} primarily include information theory-based, model-based, and machine learning-based approaches. Information theory-based methods, such as mutual information (MI) \cite{margolin2006aracne}, Pearson correlation coefficient (PCC) \cite{salleh2015reconstructing, raza2013reconstruction}, and partial information decomposition and context (PIDC) \cite{chan2017gene}, conduct correlation analyses under the assumption that the strength of the correlation between genes is is positively correlated with the likelihood of regulation between them. Model-based approaches, such as SCODE \cite{matsumoto2017scode}, involve fitting gene expression profiles to models that describe gene relationships, which are then used to reconstruct GRNs \cite{shu2021modeling, tsai2020grema}.

Machine learning-based unsupervised methods, like GENIE3 \cite{huynh2010inferring} and GRNBoost2 \cite{moerman2019grnboost2}, utilize tree-based algorithms to infer GRNs. These methods are integrated into tools like SCENIC \cite{aibar2017scenic, van2020scalable}, employing tree rules to learn regulatory relationships by iteratively excluding one gene at a time to determine its associations with other genes. Despite not requiring labeled data, these unsupervised methods often struggle with the noise, dropouts, high sparsity, and high dimensionality typical of scRNA-seq data. Additionally, the computational expense and scalability issues of these tree-based methods, due to the necessity of segmenting input data and iteratively establishing multiple models, present further challenges for large datasets.

\textbf{Supervised methods}, including DGRNS \cite{zhao2022hybrid}, convolutional neural network for co-expression (CNNC) \cite{yuan2019deep}, and DeepDRIM \cite{chen2021deepdrim}, have been developed to address the increasing scale and inherent complexity of scRNA-seq data. Compared with unsupervised learning, supervised models are capable of detecting much more subtle differences between positive and negative pairs \cite{yuan2019deep}. 

DGRNS \cite{zhao2022hybrid} combines recurrent neural networks (RNNs) for extracting temporal features and convolutional neural networks (CNNs) for extracting spatial features to infer GRNs. CNNC \cite{yuan2019deep} converts the identification of gene regulation into an image classification task by transforming the expression values of gene pairs into histograms and using a CNN for classification. However, the performance of CNNC \cite{yuan2019deep} is hindered by the issue of transitive interactions. To address this, DeepDRIM \cite{chen2021deepdrim} considers the information from neighboring genes and converts TF–gene pairs and neighboring genes into histograms as additional inputs, thereby reducing the occurrence of transitive interactions to some extent. Despite their success there exist certain limitations to the employment of CNN model-based approaches for GRN reconstruction. First of all, the generation of image data not only gives rise to unanticipated noise but also conceals certain original data features. Additionally, this process is time-consuming, and since it changes the format of scRNA-seq data, the predictions made by these CNN-based computational approaches cannot be wholly explained.

In addition to CNN-based methods, there are also other approaches such as GNE \cite{kc2019gne} and GRN-Transformer \cite{shu2022boosting}. GNE (gene network embedding) \cite{kc2019gne} is a deep learning method based on multilayer perceptron (MLP) for GRN inference applied to microarray data. It utilizes one-hot gene ID vectors from the gene topology to capture topological information, which is often inefficient due to the highly sparse nature of the resulting one-hot feature vector. GRN-Transformer \cite{shu2022boosting} constructs a weakly supervised learning framework based on axial transformer to infer cell-type-specific GRNs from scRNA-seq data and generic GRNs derived from the bulk data. 

More recently, graph neural networks (GNNs) \cite{wu2020comprehensive}, which are effective in capturing the topology of gene networks, have been introduced into GRN prediction methods. For instance, GENELink \cite{chen2022graph} treats GRN inference as a link prediction problem and uses graph attention networks to predict the probability of interaction between two gene nodes. However, existing methods often suffer from limitations such as improper handling of batch effects, difficulty in leveraging latent gene-gene interaction information, and making simplistic assumptions, which can impair their generalization and robustness. 

\section{Approach}
\begin{figure}[ht]
\vskip 0.2in
\begin{center}
\centerline{\includegraphics[width=\columnwidth]{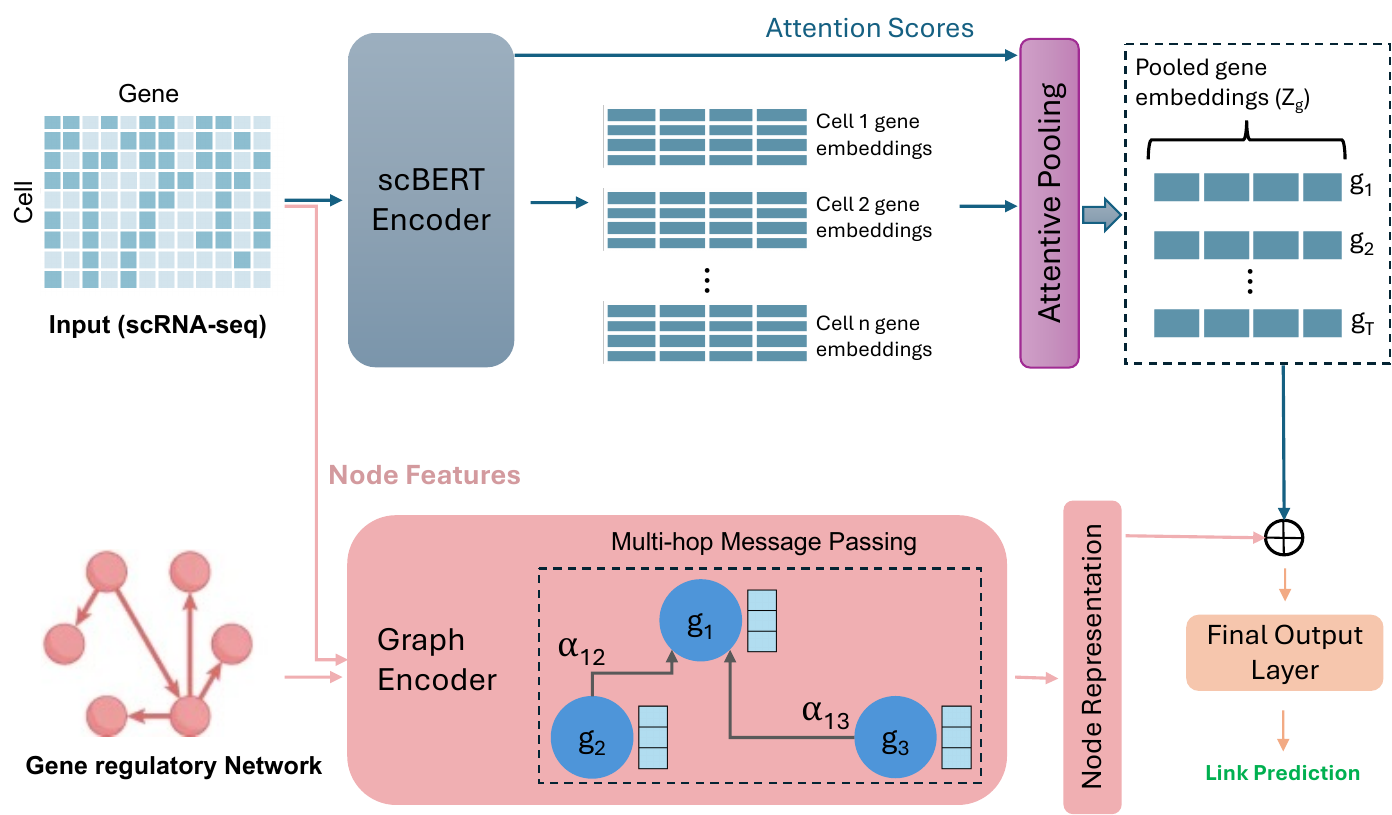}}
\caption{\textbf{Overview of scTransNet framework for supervised GRN inference} with BERT Encoding Layer (top left; \cref{module1}), Attentive Pooling (top right; \cref{module2}), GRN encoding with GNNs (bottom left; \cref{module3}) and Final Output layer (bottom right; \cref{module4}) . It augments the output from graph encoder (for knowledge understanding) with scBERT encoder (for contextual understanding) to infer regulatory interdependencies between genes.}
\label{overall_framework}
\end{center}
\vskip -0.2in
\end{figure}

As shown in (\cref{overall_framework}), our approach contains four parts: BERT encoding, Attentive Pooling, GRN encoding with GNNs and Output layer. The input scRNA-seq datasets are processed into a cell-by-gene matrix, $\textbf{X} \in R^{N \times T}$, where each element represents the read count of an RNA molecule. Specifically, for scRNA-seq data, the element denotes the RNA abundance for gene $t \in \{0, 1, ..., T\}$ in cell $n \in \{0, 1, ..., N\}$. In subsequent sections, we will refer to this matrix as the raw count matrix. Let us denote the sequence of gene tokens as $\{g_1,...,g_T\}$, where T is the total number of genes.


\subsection{BERT Encoding Layer}
\label{module1}
\cite{Yang2022, Cui2024, Chen2023, Theodoris2023} show that pre-trained transformer models have a strong understanding of gene-gene interactions across cells and have achieved state-of-the-art results on a variety of single-cell processing tasks. We use scBERT \cite{Yang2022} as the backbone, which is a successful pre-trained model with the advantage of capturing long-distance dependency as it uses Performer \cite{choromanski2022rethinking} to improve the scalability of the model to tolerate over 16,000 gene inputs.

The scBERT model adopts the advanced paradigm of BERT and tailors the architecture to solve single-cell data analysis. The connections of this model with BERT are given as follows. First, scBERT follows BERT’s revolutionary method to conduct self-supervised pre-training \cite{devlin2019bert} and uses Transformer as the model backbone \cite{choromanski2022rethinking}. Second, the design of embeddings in scBERT is similar to BERT in some aspects while having unique features to leverage gene knowledge. From this perspective, the gene expression embedding could be viewed as the token embedding of BERT. As shuffling the columns of the input does not change its meaning (like the extension of BERT to understand tabular data with TaBERT \cite{yin-etal-2020-tabert}), absolute positions are meaningless for gene. Instead gene2vec is used to produce gene embeddings, which could be viewed as relative embeddings \cite{du2019gene2vec} that capture the semantic similarities between any of two genes. Third, Transformer with global receptive field could effectively learn global representation and long-range dependency without absolute position information, achieving excellent performance on non-sequential data (such as images, tables) \cite{pmlr-v80-parmar18a, yin-etal-2020-tabert}.

In spite of the gene embedding, there is also a challenge on how to utilize the transcription level of each gene, which is actually a single continuous variable. The gene expression could also be considered as the occurrence of each gene that has already been well-documented in a biological system. Drawing from bag-of-words \cite{zhang2010understanding} insight, the conventionally used term-frequency-analysis method is applied that discretizes the continuous expression variables by binning, and converts them into 200-dimensional vectors, which are then used as token embeddings for the scBERT model.

For each token $g_t$ in a cell, we construct its input representation as:
\begin{equation}
 h_t^{0}  = emb_{gene2vec}(g_t) + emb_{expr}(g_t)
\end{equation}
where $emb_{gene2vec}(g_t)$ represents gene2vec embedding \cite{du2019gene2vec} of gene $g_t$ analogous to position embedding in BERT and $emb_{expr}(g_t)$ represents expression embedding of the gene expression of $g_t$ analogous to token embedding in BERT.

Such input representations are then fed into L successive Transformer encoder blocks, i.e.,
\begin{equation}
 h_t^{l}  = Transformer(h_t^{l-1}), l = 1, 2, ..., L,
\end{equation}
so as to generate deep, context-aware representations for genes. The final hidden states $\{h_t^{L}\}_{t=1}^{T}$ are taken as the output of this layer \cite{devlin2019bert, vaswani2023attention}.

\subsection{Attentive Pooling}
\label{module2}
After extracting the BERT encodings we further utilize the attention scores across cells from the model to select the most representative cells for pooling of each gene representation. For each input gene token $g_t$ we get the embeddings for all cells denoted as $\{h_{t(n)}^{L}\}_{n=1}^{N}$, where N is the number of cells.

The quadratic computational complexity of the BERT model, with the Transformer as its foundational unit, does not scale efficiently for long sequences. Given that the number of genes in scRNA-seq data can exceed 20,000, this limitation becomes significant. To address this issue, scBERT employs a matrix decomposition variant of the Transformer, known as Performer \cite{choromanski2022rethinking}, to handle longer sequence lengths. In a regular Transformer, the dot-product attention mechanism maps Q, K, and V, which are the encoded representations of the input queries, keys, and values for each unit. The bidirectional attention matrix is formulated as follows:
\begin{equation}
\label{attn}
    \begin{split}
        Att(Q, K, V) = D^{-1} (QK^T) V, \\
        D = diag(QK^T1_L)
    \end{split}
\end{equation}
where $Q = W_qX$, $K = W_KX$, $V = W_VX$ are linear transformations of the input X; $W_Q$, $W_K$ and $W_V$ are the weight matrices as parameters; $1_L$ is the all-ones vector of length L; and diag(.) is a diagonal matrix with the input vector as the diagonal.

The attention matrix in Performer is described as follows:
\begin{equation}
    \begin{split}
        \hat{Att}(Q, K, V) = \hat{D}^{-1}(Q^{'}((K^{'})^TV)), \\
        \hat{D} = diag(Q^{'}((K^{'})^T1_L))
    \end{split}
\end{equation}
where $Q^{'} = \phi(Q)$, $K^{'} = \phi(K)$, and the function $\phi(x)$ is defined as:
\begin{equation}
 \phi(X) = \frac{c}{\sqrt{m}}f(\omega^TX)
\end{equation}

where c is a positive constant, $\omega$ is a random feature matrix, and m is the dimensionality of the matrix. 

The attention weights can be obtained from equation \ref{attn}, modified by replacing V with $V^0$, where $V^0$ contains one-hot indicators for each position index. All the attention matrices are integrated into one matrix by taking an element-wise average across all attention matrices in multi-head multi-layer Performers. In this average attention matrix for each cell, $A (i, j)$ represents how much attention from gene i was paid to gene j. To focus on the importance of genes to each cell n, the attention matrix is summed along the columns into an attention-sum vector $a_n$, and its length is equal to the number of genes. These attention scores of gene $g_t$ are obtained across cells and normalized denoted as $\{a_t^{n}\}_{n=1}^{N}$

These normalized scores are used for weighted aggregation of gene embeddings across cells. We aggregate each cell's gene representations together into one gene-level cell embedding such that the updated matrix is of the form $Z \in R^{T \times d}$, where d is the dimension of the output gene embedding.
\begin{equation}
 Z_g[t] = \oplus_{i=1}^{N} h_{t(n)}^L \cdot a_t^n
\end{equation}

\subsection{GRN encoding with GNNs}
\label{module3}
In this module, we use raw count matrix as the features of the genes. Subsequently, we utilize graph convolutional network (GCN)-based interaction graph encoders to learn gene features by leveraging the underlying structure of the gene interaction graph.

Let us denote the prior network as $G = \{V, E\}$, where V is the set of nodes and E is the set of edges. To perform the reasoning on this prior gene regulatory network $G$, our GNN module builds on the graph attention framework (GAT) \cite{Velickovic2017GraphAN}, which induces node representations via iterative message passing between neighbors on the graph. In each layer of this GNN, the current representation of the node embeddings $\{v_1^{l-1},...,v_T^{l-1}\}$ is fed into the layer to perform a round of information propagation between nodes in the graph and yield pre-fused node embeddings for each node:
\begin{equation}
    \begin{split}
        \{\tilde{v_1}^{l},...,\tilde{v_T}^{l}\}=GNN(\{v_1^{l-1},...,v_T^{l-1}\})\\ 
        \hspace{5pt} for \hspace{5pt} l = 1,...,M
    \end{split}
\end{equation}

Specifically, for each layer l, we update the representation $\tilde{v_t}^{l}$ of each node by
\begin{equation}
     \tilde{v_t}^{l} =  f_n(\sum\limits_{v_s \in \eta_{v_t} \cup \{v_t\}} \alpha_{st} \textbf{m}_{st}) + {v_t}^{l-1}
\end{equation}
where $\eta_{v_t}$ represents the neighborhood of an arbitrary node $v_t$, $m_{st}$ denotes the message one of its neighbors $v_s$ passes to $v_t$, $\alpha_{st}$ is an attention weight that scales the message $m_{st}$, and $f_n$ is a 2-layer MLP. The messages $m_{st}$ between nodes allow entity information from a node to affect the model’s representation of its neighbors, and are computed in the following manner:
\begin{equation}
     \textbf{r}_{st} = f_r(\Tilde{\textbf{r}}_{st}, \textbf{u}_s, \textbf{u}_t)
\end{equation}
\begin{equation}
    \textbf{m}_{st} = f_m(\textbf{v}_s^{(l-1)}, \textbf{u}_s, \textbf{r}_{st})
\end{equation}

where $\textbf{u}_s$, $\textbf{u}_t$ are node type embeddings, $\Tilde{\textbf{r}}_{st}$ is a relationembedding for the relation connecting $v_s$ and $v_t$, $f_r$ is a 2-layer MLP, and $f_m$ is a linear transformation. The attention weights $\alpha_{st}$ scale the contribution of each neighbor’s message by its importance, and are computed as follows:
\begin{equation}
     \textbf{q}_s = f_q(\textbf{v}_s^{(l-1)}, \textbf{u}_s)
\end{equation}
\begin{equation}
    \textbf{k}_t = f_k(\textbf{v}_t^{(l-1)}, \textbf{u}_t, \textbf{r}_{st})
\end{equation}
\begin{equation}
    \begin{split}
        \alpha_{st} = \frac{\exp{(\gamma_{st}})}{\sum_{v_s \in \eta_{v_t} \cup \{v_t\}}\exp{(\gamma_{st}})}, 
        \gamma_{st} = \frac{\textbf{q}_s^\intercal \textbf{k}_t}{\sqrt{D}}
    \end{split}
\end{equation} 
where $f_q$ and $f_k$ are linear transformations and $\textbf{u}_s$, $\textbf{u}_t$, $\textbf{r}_{st}$ are defined the same as above.

\subsection{Final Output Layer}
\label{module4}
In the final output layer, we concatenate the input gene representations $Z_g$ from the BERT encoding layer with the graph representation of each gene from GNN to get the final gene embedding.

We input these final embeddings of pairwise genes i and j into two channels with the same structure. Each channel is composed of MLPs to further encode representations to low-dimensional vectors which serve for downstream similarity measurement or causal inference between genes.

\section{Experimental Setup}
\subsection{Benchmark scRNA-seq datasets}
The performance of scTransNet is evaluated on two human cell types using single-cell RNA-sequencing (scRNA-seq) datasets from the BEELINE study \cite{Pratapa2020}: human embryonic stem cells (hESC \cite{yuan2019deep}) and human mature hepatocytes (hHEP \cite{camp2017multilineage}). The cell-type-specific ChIP-seq ground-truth networks are used as a reference for these datasets.
The scRNA-seq datasets are preprocessed following the approach described in the \cite{Pratapa2020}, focusing on inferring interactions outgoing from transcription factors (TFs). The most significantly varying genes are selected, including all TFs with a corrected P-value (Bonferroni method) of variance below 0.01. Specifically, 500 and 1000 of the most varying genes are chosen for gene regulatory network (GRN) inference. The scRNA-seq datasets can be accessed from the Gene Expression Omnibus (GEO) with accession numbers GSE81252 (hHEP) and GSE75748 (hESC). The evaluation compares the inferred gene regulatory networks to known ChIP-seq ground-truth networks specific to these cell types.

\subsection{Implementation and Training details}
\textbf{Data Preparation}
We utilized the benchmark networks that containing labeled directed regulatory dependencies between gene pairs. These dependencies were classified as positive samples (labeled 1) if present in the network, and negative samples (labeled 0) if absent. Due to the inherent network density, the number of negative samples significantly outnumbered positive samples. To address the class imbalance, known transcription factor (TF)-gene pairs are split into training (2/3), and test (1/3) sets. Positive training samples were randomly selected from the known TF-gene pairs. Moreover, $10\%$ of TF-gene pairs are randomly selected from training samples for validation. The remaining positive pairs formed the positive test set. Negative samples were generated using the following strategies: 1) Unlabeled interactions: All unobserved TF-gene interactions outside the labeled files were considered negative instances. 2) Hard negative sampling: To enhance model learning during training, we employed a uniformly random negative sampling strategy within the training set. This involved creating "hard negative samples" by pairing each positive sample (g1, g2) with a negative sample (g1, g3), where both share the same gene g1. This approach injects more discriminative information and accelerates training. 3) Information leakage prevention: Negative test samples were randomly selected from the remaining negative instances after generating the training and validation sets. This ensured no information leakage from the test set to the training process. The positive-to-negative sample ratio in each dataset was adjusted to reflect the network density i.e.

\begin{equation}
    \frac{Positive}{Negative} = \frac{NetworkDensity}{1-NetworkDensity}
\end{equation}

\textbf{Model Training}
To account for the class imbalance, we adopted two performance metrics: Area Under the Receiver Operating Characteristic Curve (AUROC) and Area Under the Precision-Recall Curve (AUPRC). The supervised model was trained for 100 iterations with a learning rate of 0.003. The Graph Neural Network (GNN) architecture comprised two layers with hidden layer sizes of 256 and 128 units, respectively.

\textbf{Evaluation}
All reported results are based solely on predictions from the held-out test set. To ensure a fair comparison, identical training and validation sets were utilized for all evaluated supervised methods. This approach eliminates potential bias introduced by different data splits.

\begin{figure*}[ht]
\vskip 0.2in
\begin{center}
\centerline{\includegraphics[width=\textwidth]{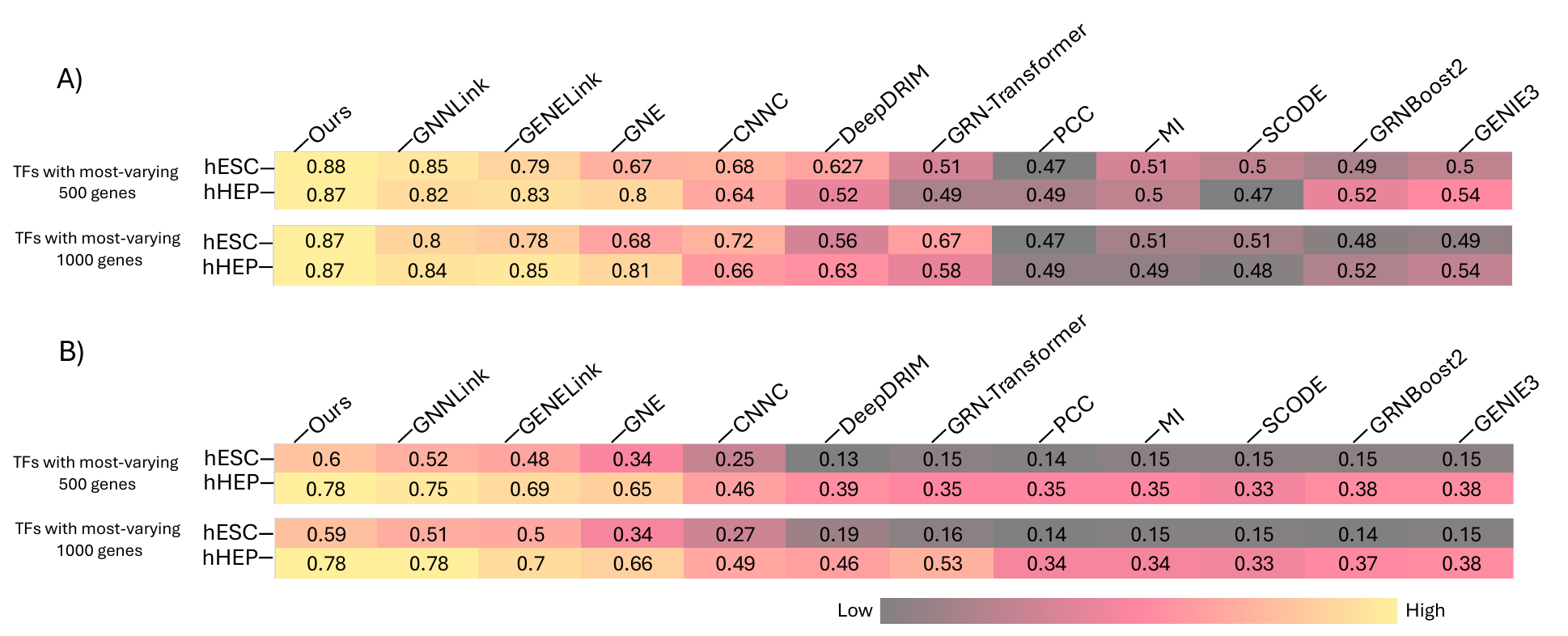}}
\caption{Summary of the GRN prediction performance of scTransNet in the (A) AUROC metric (top) (B) and the AUPRC metric (bottom). Our evaluation is conducted on two human single-cell RNA sequencing (scRNA-seq) datasets, with a cell-type-specific ground-truth network. The scRNA-seq datasets consist of significantly varying transcription factors (TFs) and the 500 or 1000 most-varying genes.}
\label{result}
\end{center}
\vskip -0.2in
\end{figure*}

\subsection{Baseline Methods}
To assess the effectiveness of our model in predicting GRNs, we compare our model scTransNet against the existing baseline methods commonly used for inferring GRNs, as follows:

\begin{itemize}[leftmargin=*]
  \item GNNLink \cite{10.1093/bib/bbad414} is a graph neural network model that uses a GCN-based interaction graph encoder to capture gene expression patterns.
  \item GENELink \cite{chen2022graph} proposes a graph attention network (GAT) approach to infer potential GRNs by leveraging the graph structure of gene regulatory interactions.
  \item GNE (gene network embedding) \cite{kc2019gne} proposes a multilayer perceptron (MLP) approach to encode both gene expression profiles and network topology for predicting gene dependencies.
  \item CNNC \cite{yuan2019deep} proposes inferring GRNs using deep convolutional neural networks (CNNs).
  \item DeepDRIM \cite{chen2021deepdrim} is a supervised deep neural network that utilizes images representing the expression distribution of joint gene pairs as input for binary classification of regulatory relationships, considering both target TF-gene pairs and potential neighbor genes.
  \item GRN-transformer \cite{shu2022boosting} is a weakly supervised learning method that utilizes axial transformers to infer cell type-specific GRNs from single-cell RNA-seq data and generic GRNs.
  \item Pearson correlation coefficient (PCC) \cite{salleh2015reconstructing, raza2013reconstruction} is a traditional statistical method for measuring the linear correlation between two variables, often used as a baseline for GRN inference.
  \item Mutual information (MI) \cite{margolin2006aracne} is an information-theoretic measure of the mutual dependence between two random variables, also used as a baseline for GRN inference.
  \item SCODE \cite{matsumoto2017scode} is a computational method for inferring GRNs from single-cell RNA-seq data using a Bayesian framework.
  \item GRNBoost2 \cite{moerman2019grnboost2} is a gradient boosting-based method for GRN inference.
  \item GENIE3 \cite{huynh2010inferring} is a random forest-based machine learning method that constructs GRNs based on regression weight coefficients, and won the DREAM5 In Silico Network Challenge in 2010.
\end{itemize}

These methods represent a diverse range of approaches, including traditional statistical methods, machine learning techniques, and deep learning models, for inferring gene regulatory networks from various types of data, such as bulk and single-cell RNA-seq, as well as incorporating additional information like network topology and chromatin accessibility.

\begin{table*}
\caption{Comparison of average AUROC and AUPRC evaluation metrics on human benchmark datasets, validating the roles of the GNN encoder, scBERT encoder, and Attentive Pooling using the cell-type-specific ChIP-seq network for our proposed method.}
\label{ablation_modules}
\vskip 0.15in
\begin{center}
\begin{small}
\begin{tabular}{lcccccccc} 
\toprule
& \multicolumn{2}{c}{\textbf{w/o GNN encoder}} & \multicolumn{2}{c}{\textbf{w/o scBERT encoder}} & \multicolumn{2}{c}{\textbf{w/o Attentive Pooling}} & \multicolumn{2}{c}{\textbf{scTransNet}} \\
\cmidrule(lr){2-3} \cmidrule(lr){4-5} \cmidrule(lr){6-7} \cmidrule(lr){8-9}
\textbf{Dataset} & \textbf{AUROC} & \textbf{AUPRC} & \textbf{AUROC} & \textbf{AUPRC} & \textbf{AUROC} & \textbf{AUPRC} & \textbf{AUROC} & \textbf{AUPRC}\\
\midrule
hESC    & 0.842& 0.544& 0.853& 0.572& 0.860& 0.569& \textbf{0.880}& \textbf{0.595}\\
hHEP & 0.830& 0.725& 0.854& 0.753& 0.862& 0.683& \textbf{0.870}& \textbf{0.780}\\
\bottomrule
\end{tabular}
\end{small}
\end{center}
\vskip -0.1in
\end{table*}

\section{Results}

\subsection{Performance on benchmark datasets}
The results (see \cref{result}) demonstrate that scTransNet outperforms state-of-the-art baseline methods across all four benchmark datasets, achieving superior performance in terms of both AUROC and AUPRC evaluation metrics. Notably, scTransNet's AUROC values are approximately $5.4\%$ and $7.4\%$ higher on average compared to the second-best methods, namely GNNLink \cite{10.1093/bib/bbad414} and GENELink \cite{chen2022graph}, respectively. Similarly, scTransNet's AUPRC values show an impressive improvement of approximately $7.4\%$ and $16\%$ on average over GNNLink and GENELink, respectively.

To gain further insights, we analyzed scTransNet's final gene regulatory network (GRN) predictions and compared them with those from GENELink. Our analysis revealed that scTransNet effectively captured all the gene regulatory interactions predicted by GENELink. This finding suggests that by incorporating joint learning, scTransNet does not introduce additional noise to the predictive power of the graph representations. Instead, it enhances the predictive capability through the scBERT encoder in its architecture.

\begin{figure}[ht]
\vskip 0.2in
\begin{center}
\centerline{\includegraphics[width=\columnwidth]{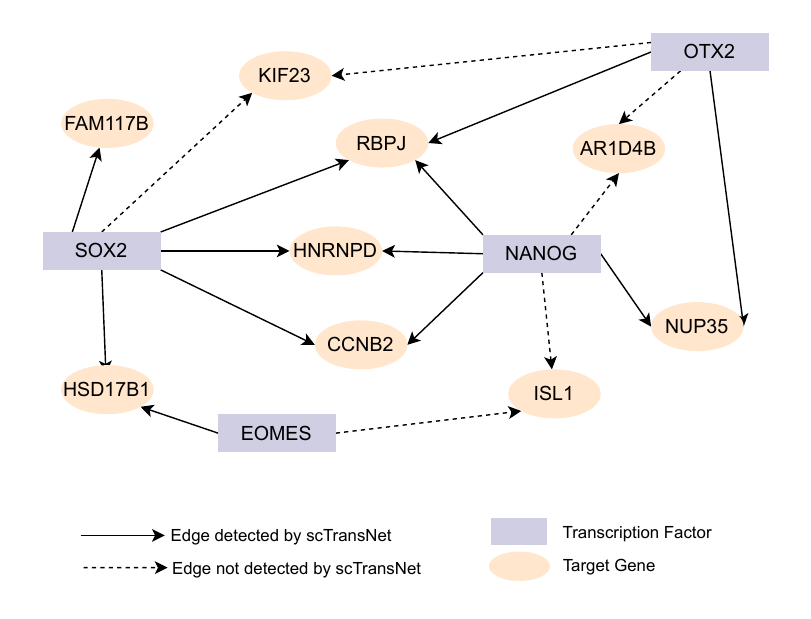}}
\caption{GRN prediction performance of scTransNet on a partial ground truth subgraph. Solid line edges depict ground truth regulatory interactions correctly predicted by scTransNet but missed by the baseline GENELink method, which relies solely on graph representations. Notably, scTransNet effectively identified all regulatory links predicted by GENELink (not visualized). Dotted line edges represent ground truth interactions that scTransNet failed to capture reveal its limitations and providing insights for further improvement. Overall, this highlights scTransNet's strength in leveraging joint learning to uncover additional true regulatory interactions beyond graphs.}
\label{GRN_visualization}
\end{center}
\vskip -0.2in
\end{figure}

\cref{GRN_visualization} provides a visualization of a partial subgraph of the ground truth GRN, highlighting the predictions made by scTransNet that were not captured by GENELink, which solely relies on graphs for predicting gene-gene interactions. Additionally, the figure visualizes the ground truth labels that scTransNet failed to capture. In summary, the comparative analysis demonstrates that scTransNet effectively captures all the regulatory interactions predicted by GENELink while leveraging joint learning to improve predictive performance. The visualization illustrates the additional interactions scTransNet could predict beyond GENELink, as well as the ground truth interactions it missed, providing insights into the strengths and limitations of the proposed method.

\subsection{Discussion and Ablations}
To evaluate the effectiveness of jointly learning from pre-trained scRNA-seq language models \cite{Yang2022}, which capture rich contextual representations, and Gene Regulatory Networks (GRNs), which encode structured biological knowledge, we compare the average Area Under the Receiver Operating Characteristic Curve (AUROC) and Area Under the Precision-Recall Curve (AUPRC) metrics with and without these encoders across the four human cell type benchmark datasets \cite{Pratapa2020}. The average AUROC and AUPRC scores are calculated across both the TFs+500 highly variable genes and TFs+1000 highly variable genes datasets for each human cell data type (i.e, hESC (human embryonic stem cells) and hHEP (human mature hepatocytes)). Additionally, we validate the importance of incorporating Attentive Pooling (\cref{module2}) by contrasting the results when using average pooling of gene embeddings across cells instead of attentive pooling. Consistent parameter settings are employed across all four human cell benchmark datasets, with Cell-type-specific ChIP-seq network data serving as the ground truth.

\textbf{Effect of Graph Neural Network Component:} The results demonstrate the significant impact of incorporating the Graph Neural Network (GNN) encoder component in the proposed method. With the GNN encoder, the average AUROC value across all the human cell type datasets is $87.5\%$, and the average AUPRC value is $68.7\%$. In contrast, without the GNN encoder, the average AUROC drops to $83.6\%$, and the average AUPRC decreases to $63.4\%$. The inclusion of the GNN encoder leads to an improvement of $4.6\%$ in the average AUROC and a notable $8.3\%$ increase in the average AUPRC. These results highlight the consistent performance enhancement provided by the GNN encoder across both AUROC and AUPRC metrics for the human cell type benchmark datasets. The GNN encoder plays a crucial role in the architecture as the task is formulated as a supervised Gene Regulatory Network (GRN) inference problem, aiming to identify potential gene regulatory dependencies given prior knowledge of the GRN. The GNN models the regulatory interactions as a graph, learning node representations that effectively encode the network topology and gene interdependencies present in the GRN, which serves as the primary source of biological knowledge.  The results in \cref{ablation_modules} justify the use of this structural graph representation for understanding the complex regulatory networks in single-cell transcriptomics data.

\textbf{Effect of Pre-trained Single-Cell Transcriptomics Transformer:} The removal of the scBERT encoder also leads to a drop in performance, with the average AUROC decreasing from $87.5\%$ to $85.3\%$, and the average AUPRC declining from $68.7\%$ to $66.2\%$ across both cell types (see \cref{ablation_modules}). The inclusion of scBERT representations improves the AUROC by $2.6\%$ and the AUPRC by $3.8\%$. While the improvement is less significant compared to the GNN encoder, this is expected as the contextual representations from scRNA-seq data are learned through pre-training on millions of unlabeled single cells and then fine-tuned for the specific cell type. In addition to rich contextual representations, scBERT captures long-range dependencies between genes by leveraging self-attention mechanisms and pretraining on large-scale unlabeled scRNA-seq data \cite{Pratapa2020}. This comprehensive understanding of gene-gene interactions and semantic relationships allows for effective modeling of complex, non-linear gene regulatory patterns that extend beyond immediate neighbors in the gene regulatory network.

The contextual representations learned by the pre-trained Transformer facilitate the identification of intricate regulatory relationships that might be overlooked by traditional methods focused on local neighborhoods or predefined gene sets. The ability to capture global context and long-range dependencies is a key advantage of pre-trained single-cell Transformer models for deciphering the intricate gene regulatory mechanisms governing cellular states and identities. The improvement shown in Table 1 justifies the effectiveness of this approach.

\textbf{Effect of Attentive Pooling Mechanism:} The impact of incorporating Attentive Pooling is evaluated by comparing the AUROC and AUPRC metrics with and without attentive pooling across four datasets. As shown in \cref{ablation_modules}, the inclusion of attentive pooling results in a slight improvement, with a $1.6\%$ increase in the average AUROC and a $9.6\%$ increase in the average AUPRC. While the improvement is not significant, the experiments confirm that attentive pooling offers some support for the gene regulation task. We believe that the significance of attentive pooling will be more pronounced when scaling the method to larger datasets. The cell type data is sparse and of low quality. However, the attention weights learned from scBERT \cite{Pratapa2020} demonstrate that the marker genes are automatically learned for each cell. Consequently, attentive pooling helps to effectively focus on high-quality cell data by removing noise. By employing an attentive pooling mechanism, scTransNet selectively focuses on the most informative cells for each gene, mitigating noise and filtering out irrelevant information, thereby enhancing the quality of the input data used for GRN inference.

\section{Conclusion and Future Work}
In this work, we propose scTransNet, a joint graph learning inference framework that integrates prior knowledge from known Gene Regulatory Networks (GRNs) with contextual representations learned by pre-trained single-cell transcriptomics Transformers. Our approach aims to effectively boost GRN prediction by leveraging the complementary strengths of structured biological knowledge and rich contextual representations. We evaluate our method on four human cell scRNA-seq benchmark datasets and demonstrate consistent improvements over current baselines in predicting gene-gene regulatory interactions. Our framework comprises four key modules: a GNN encoder to capture the network topology from known GRNs, a scBERT encoder to learn contextual representations from scRNA-seq data, an Attentive Pooling mechanism to focus on informative cells, and a Final Output layer for prediction. The synergistic combination of these modules is verified to be effective in accurately inferring gene regulatory dependencies.

Moving forward, we plan to incorporate the knowledge integration process directly into the fine-tuning of the Transformer model, aiming to fuse information across layers more effectively. Additionally, we will evaluate our approach on various other datasets, including simulated datasets, to further validate its robustness and generalizability. Beyond GRN inference, we intend to investigate the advantages of jointly learning single-cell Transformers and structured biological knowledge for other cell-related tasks. These tasks include cell type annotation, identifying echo archetypes \cite{LUCA20215482}, and enhancing the interpretability of single-cell models. By leveraging the complementary strengths of contextual representations and structured knowledge, we aim to advance the understanding and analysis of complex cellular processes and regulatory mechanisms.

\section*{Acknowledgements}
Our work is sponsored by the NSF NAIRR Pilot and PSC Neocortex, Commonwealth Cyber Initiative, Children’s National Hospital, Fralin Biomedical Research Institute (Virginia Tech), Sanghani Center for AI and Data Analytics (Virginia Tech), Virginia Tech Innovation Campus, and a generous gift from the Amazon + Virginia Tech Center for Efficient and Robust Machine Learning.
\section*{Impact Statement}

``This paper presents work whose goal is to advance the field of 
Machine Learning. There are many potential societal consequences 
of our work, none which we feel must be specifically highlighted here.''



\bibliography{example_paper}
\bibliographystyle{icml2024}




\end{document}